\documentclass{article}

     \PassOptionsToPackage{numbers, compress}{natbib}

\usepackage[final]{neurips_2020}

\usepackage[utf8]{inputenc} 
\usepackage[T1]{fontenc}    
\usepackage{microtype}
\usepackage{graphicx}
\usepackage{subfigure}
\usepackage{booktabs} 
\usepackage{times}
\usepackage{epsfig}
\usepackage{amsmath}
\usepackage{bm}
\usepackage{amsthm}
\usepackage{multirow}
\usepackage[super]{nth}

\usepackage{amssymb}%
\usepackage{mathrsfs}%
\usepackage{nicefrac}
\usepackage{color,xcolor}
\usepackage{pifont}
\usepackage{array}
\usepackage{hyperref}
\usepackage{caption}
\usepackage{wrapfig}

\definecolor{myblueblue}{RGB}{238 18 137}

\newtheorem{lemma}{Lemma}

\newtheorem{lemmaappendix}{Lemma}

\DeclareMathOperator*{\argmin}{arg\,min}
\DeclareMathOperator*{\argmax}{arg\,max}

\DeclareMathOperator{\sgn}{sign}
\DeclareMathOperator{\softmax}{\mathbb{S}}
\DeclareMathOperator{\R}{\mathbb{R}}

\DeclareMathOperator{\Loss}{\mathcal{L}}
\DeclareMathOperator{\CE}{\text{CE}}

\hypersetup{
	colorlinks=true,
	urlcolor=magenta,
}

\title{Boosting Adversarial Training with\\ Hypersphere Embedding}

\renewcommand\footnotemark{}
\author{
Tianyu Pang$^*$, Xiao Yang$^*$, Yinpeng Dong, Kun Xu, Jun Zhu, Hang Su$^\dagger$ \thanks{$^*$Equal contribution. $^\dagger$  Corresponding author.}\\
  Dept. of Comp. Sci. \& Tech., Institute for AI, BNRist Center\\
Tsinghua-Bosch Joint ML Center, THBI Lab, Tsinghua University, Beijing, China\\
  \texttt{\footnotesize \{pty17, yangxiao19, dyp17\}@mails.tsinghua.edu.cn}\\
  \texttt{\footnotesize kunxu.thu@gmail.com, \{suhangss, dcszj\}@mail.tsinghua.edu.cn} \\
}

\begin{document}

\maketitle

\begin{abstract}
Adversarial training (AT) is one of the most effective defenses against adversarial attacks for deep learning models. In this work, we advocate incorporating the hypersphere embedding (HE) mechanism into the AT procedure by regularizing the features onto compact manifolds, which constitutes a lightweight yet effective module to blend in the strength of representation learning. Our extensive analyses reveal that AT and HE are well coupled to benefit the robustness of the adversarially trained models from several aspects. We validate the effectiveness and adaptability of HE by embedding it into the popular AT frameworks including PGD-AT, ALP, and TRADES, as well as the FreeAT and FastAT strategies. In the experiments, we evaluate our methods under a wide range of adversarial attacks on the CIFAR-10 and ImageNet datasets, which verifies that integrating HE can consistently enhance the model robustness for each AT framework with little extra computation.
\end{abstract}

\section{Introduction}
The adversarial vulnerability of deep learning models has been widely recognized in recent years~\citep{biggio2013evasion,Goodfellow2014,Szegedy2013}. To mitigate this potential threat, a number of defenses have been proposed, but most of them are ultimately defeated by the attacks adapted to the specific details of the defenses~\citep{athalye2018obfuscated,carlini2017adversarial}. Among the existing defenses, \textbf{adversarial training (AT)} is a general strategy achieving the state-of-the-art robustness under different settings~\citep{salman2019provably,tramer2017ensemble,wu2019defending,xie2018feature,xu2020dynamic,yang2019improving,zhu2019freelb}. Various efforts have been devoted to improving AT from different aspects, including accelerating the training procedure~\citep{schwinn2020fast,shafahi2019adversarial,wong2020fast,zhang2019you} and exploiting extra labeled and unlabeled training data~\citep{alayrac2019labels,carmon2019unlabeled,hendrycks2019using,zhai2019adversarially}, which are conducive in the cases with limited computational resources or additional data accessibility.

In the meanwhile, another research route focuses on boosting the adversarially trained models via imposing more direct supervision to regularize the learned representations. Along this line, recent progress shows that encoding triplet-wise metric learning or maximizing the optimal transport (OT) distance of data batch in AT is effective to leverage the inter-sample interactions, which can promote the learning of robust classifiers~\citep{li2019improving,mao2019metric,pang2019improving,zhang2019defense}. However, optimization on the sampled triplets or the OT distance is usually of high computational cost, while the sampling process in metric learning could also introduce extra class biases on unbalanced data~\citep{parkhi2015deep,schroff2015facenet}.

In this work, we provide a lightweight yet competent module to tackle several defects in the learning dynamics of existing AT frameworks, and facilitate the adversarially trained networks learning more robust features. Methodologically, we augment the AT frameworks by integrating the \textbf{hypersphere embedding (HE)} mechanism, which normalizes the features in the penultimate layer and the weights in the softmax layer with an additive angular margin. Except for the generic benefits of HE on learning angularly discriminative representations~\citep{liu2017sphereface,liu2017deepdefense,wang2017normface,wu2018unsupervised}, we contribute to the extensive analyses (detailed in Sec.~\ref{analysis}) showing that the encoded HE mechanism naturally adapts to AT.

To intuitively explain the main insights, we take a binary classification task as an example, where the cross-entropy (CE) objective equals to maximizing $\mathcal{L}(x)=(W_{0}-W_{1})^{\top}z=\|W_{01}\|\|z\|\cos(\theta)$ on an input $x$ with label $y=0$. (\textbf{\romannumeral 1}) If $x$ is correctly classified, there is $\mathcal{L}(x)>0$, and adversaries aim to craft $x'$ such that $\mathcal{L}(x')<0$. Since $\|W_{01}\|$ and $\|z\|$ are always positive, they cannot alter the sign of $\mathcal{L}$. Thus feature normalization (FN) and weight normalization (WN) encourage the adversaries to attack the crucial component $\cos(\theta)$, which results in more efficient perturbations when crafting adversarial examples in AT; (\textbf{\romannumeral 2}) In a data batch, points with larger $\|z\|$ will dominate (vicious circle on increasing $\|z\|$), which makes the model ignore the critical component $\cos(\theta)$. FN alleviates this problem by encouraging the model to devote more efforts on learning hard examples, and well-learned hard examples will dynamically have smaller weights during training since $\cos(\theta)$ is bounded; This can promote the worst-case performance under adversarial attacks; (\textbf{\romannumeral 3}) When there are much more samples of label $0$, the CE objective will tend to have $\|W_{0}\|\gg\|W_{1}\|$ to minimize the loss. WN can relieve this trend and encourage $W_{0}$ and $W_{1}$ to diversify in directions. This mechanism alleviates the unbalanced label distributions caused by the untargeted or multi-targeted attacks applied in AT~\citep{gowal2019alternative,madry2018towards}, where the resulted adversarial labels depend on the semantic similarity among classes; (\textbf{\romannumeral 4}) The angular margin (AM) induces a larger inter-class variance and margin under the angular metric to further improve model robustness, which plays a similar role as the margin in SVM.

{
\renewcommand*{\arraystretch}{1.6}
\begin{table*}[t]
\vspace{-0.2cm}
  \caption{Formulations of the AT frameworks without (\ding{54}) or with (\ding{52}) HE. The notations are defined in Sec.~\ref{HE in face}. We substitute the adversarial attacks in ALP with {untargeted PGD} as suggested~\citep{engstrom2018evaluating}.}
  \vspace{-0.3cm}
  \begin{center}
  \begin{small}
  \begin{tabular}{c|c|c|c}
\hline
\!\!Strategy\! &  \!HE\! & Training objective $\Loss_{\text{T}}$  & Adversarial objective $\Loss_{\text{A}}$ \\

\hline

\!\!\multirow{2}{*}{PGD-AT}\!&\ding{54}&$\Loss_{\CE}(f(x^*),y)$ & $\Loss_{\CE}(f(x'),y)$\\

\cline{2-4}

&\ding{52}&
$\Loss_{\CE}^{{\color{myblueblue}m}}({\color{myblueblue}\widetilde{f}}(x^{*}),y)$ & $\Loss_{\CE}({\color{myblueblue}\widetilde{f}}(x'),y)$\\

\hline

\!\!\multirow{2}{*}{ALP}\!&\ding{54}& $\alpha\Loss_{\CE}({f}(x),y)\!+\!(1\!-\!\alpha)\Loss_{\CE}({f}(x^{*}),y)\!+\!\lambda\|\mathbf{W}^{\top}\!(z\!-\!z^{*})\|_{2}$ & $\Loss_{\CE}(f(x'),y)$\\
  
\cline{2-4}
  
&\ding{52}& $\alpha\Loss_{\CE}^{{\color{myblueblue}m}}({\color{myblueblue}\widetilde{f}}(x),y)\!+\!(1\!-\!\alpha)\Loss_{\CE}^{{\color{myblueblue}m}}({\color{myblueblue}\widetilde{f}}(x^{*}),y)\!+\!\lambda\|{\color{myblueblue}\widetilde{\mathbf{W}}}^{\top}\!({\color{myblueblue}\widetilde{z}}\!-\!{\color{myblueblue}\widetilde{z}^{*}})\|_{2}$ & $\Loss_{\CE}({\color{myblueblue}\widetilde{f}}(x'),y)$\\

\hline

\!\!\multirow{2}{*}{TRADES}\!&\ding{54}& $\Loss_{\CE}({f}(x),y)+\lambda\Loss_{\CE}({f}(x^{*}),f(x))$ & $\Loss_{\CE}(f(x'),f(x))$\\
  
\cline{2-4}
  
&\ding{52}& $\Loss_{\CE}^{{\color{myblueblue}m}}({\color{myblueblue}\widetilde{f}}(x),y)+\lambda\Loss_{\CE}({\color{myblueblue}\widetilde{f}}(x^{*}),{\color{myblueblue}\widetilde{f}}(x))$ & $\Loss_{\CE}({\color{myblueblue}\widetilde{f}}(x'),{\color{myblueblue}\widetilde{f}}(x))$\\

\hline

   \end{tabular}
  \end{small}
  \end{center}
  \label{table:6}
  \vspace{-0.2cm}
\end{table*}
}

Our method is concise and easy to implement. To validate the effectiveness, we consider three typical AT frameworks to incorporate with HE, namely, \textbf{PGD-AT}~\citep{madry2018towards}, \textbf{ALP}~\citep{kannan2018adversarial}, and \textbf{TRADES}~\citep{zhang2019theoretically}, as summarized in Table~\ref{table:6}. We further verify the generality of our method by evaluating the combination of HE with previous strategies on accelerating AT, e.g., \textbf{FreeAT}~\citep{shafahi2019adversarial} and \textbf{FastAT}~\citep{wong2020fast}. In Sec.~\ref{Experiements}, we empirically evaluate the defenses on CIFAR-10~\citep{Krizhevsky2012} and ImageNet~\citep{deng2009imagenet} under several different adversarial attacks, including the commonly adopted PGD~\citep{madry2018towards} and other strong ones like the feature attack~\citep{featureattack}, FAB~\citep{croce2019minimally}, SPSA~\citep{uesato2018adversarial}, and NES~\citep{ilyas2018black}, etc. We also test on the CIFAR-10-C and ImageNet-C datasets with corrupted images to inspect the robustness under general transformations~\citep{hendrycks2019benchmarking}. The results demonstrate that incorporating HE can consistently improve the performance of the models trained by each AT framework, while introducing little extra computation.

\vspace{-0.05cm}
\section{Methodology}
\vspace{-0.1cm}
\label{methodology}
In this section, we define the notations, introduce the hypersphere embedding (HE) mechanism, and provide the formulations under the adversarial training (AT) frameworks. Due to the limited space, we extensively introduce the related work in Appendix {\color{red}B}, including those on combining metric learning with AT~\citep{li2019improving,mao2019metric,pang2018max,pang2019rethinking,zhang2019defense} and further present their bottlenecks.

\vspace{-0.05cm}
\subsection{Notations}
\vspace{-0.1cm}
For the classification task with $L$ labels in $[L]:=\{1,\cdots,L\}$, a deep neural network (DNN) can be generally denoted as the mapping function $f(x)$ for the input $x$ as
\begin{equation}
    f(x)=\softmax(\mathbf{W}^{\top}z+b)\text{,}
    \label{equ:1}
\end{equation}
where $z=z(x;\bm{\omega})$ is the extracted feature with model parameters $\bm{\omega}$, the matrix $\mathbf{W}=(W_{1},\cdots,W_{L})$ and vector $b$ are respectively the weight and bias in the softmax layer, and $\softmax(h):\R^{L}\rightarrow\R^{L}$ is the softmax function. One common training objective for DNNs is the cross-entropy (CE) loss defined as
\begin{equation}
    \Loss_{\CE}(f(x),y)=-1_{y}^{\top}\log f(x),
    \label{equ:2}
\end{equation}
where $1_{y}$ is the one-hot encoding of label $y$ and the logarithm of a vector is taken element-wisely. In this paper, we use $\angle(u,v)$ to denote the angle between vectors $u$ and $v$.

\vspace{-0.05cm}
\subsection{The AT frameworks with HE}
\vspace{-0.1cm}
\label{HE in face}
Adversarial training (AT) is one of the most effective and widely studied defense strategies against adversarial vulnerability~\citep{brendel2020adversarial,kurakin2018competation}. Most of the AT methods can be formulated as a two-stage framework:
\begin{equation}
    \min_{\bm{\omega},\mathbf{W}}\mathbb{E}\left[\Loss_{\text{T}}(\bm{\omega},\mathbf{W}|x,x^*,y)\right]\text{, where }x^{*}=\argmax_{x'\in\mathbf{B}(x)}\Loss_{\text{A}}(x'|x,y,\bm{\omega},\mathbf{W})\text{.}
\end{equation}
Here $\mathbb{E}[\cdot]$ is the expectation w.r.t. the data distribution, $\mathbf{B}(x)$ is a set of allowed points around $x$, $\Loss_{\text{T}}$ and $\Loss_{\text{A}}$ are the training and adversarial objectives, respectively. Since the inner maximization and outer minimization problems are mutually coupled, they are iteratively executed in training until the model parameters $\bm{\omega}$ and $\mathbf{W}$ converge~\citep{madry2018towards}. To promote the performance of the adversarially trained models, recent work proposes to embed pair-wise or triplet-wise metric learning into AT~\citep{li2019improving,mao2019metric,zhang2019defense}, which facilitates the neural networks learning more robust representations. Although these methods are appealing, they could introduce high computational overhead~\citep{mao2019metric}, cause unexpected class biases~\citep{hoffer2015deep}, or be vulnerable under strong adversarial attacks~\citep{featureattack}.


In this paper, we address the above deficiencies by presenting a lightweight yet effective module that integrates the \textbf{hypersphere embedding (HE)} mechanism with an AT procedure. Though HE is not completely new, our analysis in Sec.~\ref{analysis} demonstrates that HE naturally adapts to the learning dynamics of AT and can induce several advantages special to the adversarial setting. Specifically, the HE mechanism involves three typical operations including feature normalization (FN), weight normalization (WN), and angular margins (AM), as described below.

Note that in Eq.~(\ref{equ:1}) there is $\mathbf{W}^{\top}z=(W_{1}^{\top}z,\cdots,W_{L}^{\top}z)$, and $\forall l\in[L]$, the inner product $W_{l}^{\top}z=\|W_{l}\|\|z\|\cos{(\theta_{l})}$, where $\theta_{l}=\angle(W_{l},z)$.\footnote{We omit the subscript of $\ell_2$-norm without ambiguity.} Then the WN and FN operations can be denoted as
\begin{equation}
\textbf{WN operation: }\widetilde{W}_{l}=\frac{W_{l}}{\|W_{l}\|}\text{; }\textbf{FN operation: }\widetilde{z}=\frac{z}{\|z\|}\text{,}
\end{equation}
Let $\cos\bm{\theta}=(\cos{(\theta_{1})},\cdots,\cos{(\theta_{L})})$ and $\widetilde{\mathbf{W}}$ be the weight matrix after executing WN on each column vector $W_{l}$. Then, the output predictions of the DNNs with HE become
\begin{equation}
     \widetilde{f}(x)=\softmax(\widetilde{\mathbf{W}}^{\top}\widetilde{z})=\softmax(\cos{\bm{\theta}})\text{,}
     \label{equ:4}
\end{equation}
where no bias vector $b$ exists in $\widetilde{f}(x)$~\cite{liu2017deep,wang2017normface}. In contrast, the \textbf{AM operation} is only performed in the training phase, where $\widetilde{f}(x)$ is fed into the CE loss with a margin $m$~\citep{wang2018cosface}, formulated as 
\begin{equation}
    \Loss_{\CE}^{m}(\widetilde{f}(x),y)=-1_{y}^{\top}\log\softmax(s\cdot (\cos\bm{\theta}-m\cdot 1_{y}))\text{.}
    \label{CE_HE}
\end{equation}
Here $s>0$ is a hyperparameter to improve the numerical stability during training~\cite{wang2017normface}. To highlight our main contributions in terms of methodology, we summarize the proposed formulas of AT in Table~\ref{table:6}. We mainly consider three representative AT frameworks including \textbf{PGD-AT}~\citep{madry2018towards}, \textbf{ALP}~\citep{kannan2018adversarial}, and \textbf{TRADES} \citep{zhang2019theoretically}. The differences between our enhanced versions (with HE) from the original versions (without HE) are colorized. Note that we apply the HE mechanism both on the adversarial objective $\Loss_{\text{A}}$ for constructing adversarial examples (the inner maximization problem), and the training objective $\Loss_{\text{T}}$ for updating parameters (the outer minimization problem).


\section{Analysis of the benefits}
\label{analysis}
In this section, we analyze the benefits induced by the mutual interaction between AT and HE under the $\ell_{p}$-bounded threat model~\citep{carlini2019evaluating}, where $\mathbf{B}(x)=\{x'|\|x'-x\|_{p}\leq\epsilon\}$ and $\epsilon$ is the maximal perturbation. Detailed proofs for the conclusions below can be found in Appendix {\color{red} A}.

\subsection{Formalized first-order adversary}
Most of the adversarial attacks applied in AT belong to the family of first-order adversaries~\citep{simon2019first}, due to the computational efficiency.
We first define the vector function $\mathbb{U}_{p}$ as
\begin{equation}
    \mathbb{U}_{p}(u)=\argmax_{\|v\|_{p}\leq 1}u^{\top}v\text{, where }u^{\top}\mathbb{U}_{p}(u)=\|u\|_{q}\text{.}
    \label{Up}
\end{equation}
Here $\|\cdot\|_{q}$ is the dual norm of $\|\cdot\|_{p}$ with $\frac{1}{p}+\frac{1}{q}=1$~\citep{boyd2004convex}. Specially, there are $\mathbb{U}_{2}(u)=\frac{u}{\|u\|_{2}}$ and $\mathbb{U}_{\infty}(u)=\sgn(u)$. If $u=\nabla_{x}\Loss_{\text{A}}$, then $\mathbb{U}_{p}(\nabla_{x}\Loss_{\text{A}})$ is the direction of greatest increase of $\Loss_{\text{A}}$ under the first-order Taylor's expansion~\citep{konigsberger2004analysis} and the $\ell_{p}$-norm constraint, as stated below:
\begin{lemma}
\label{lemma1}
(First-order adversary) Given the adversarial objective $\Loss_{\text{A}}$ and the set $\mathbf{B}(x)=\{x'|\|x'-x\|_{p}\leq\epsilon\}$, under the first-order Taylor's expansion, the solution for $\max_{x'\in{B}(x)}\Loss_{\text{A}}(x')$ is $x^*=x+\epsilon\mathbb{U}_{p}(\nabla_{x}\Loss_{\text{A}}(x))$. Furthermore, there is $\Loss_{\text{A}}(x^*)=\Loss_{\text{A}}(x)+\epsilon\|\nabla_{x}\Loss_{\text{A}}(x)\|_{q}$. 
\end{lemma}
According to the one-step formula in Lemma~\ref{lemma1}, we can generalize to the multi-step generation process of first-order adversaries under the $\ell_{p}$-bounded threat model. For example, in the $t$-th step of the iterative attack with step size $\eta$~\citep{kurakin2016}, the adversarial example $x^{(t)}$ is updated as
\begin{equation}
    x^{(t)}=x^{(t-1)}+\eta\mathbb{U}_{p}(\nabla_{x}\Loss_{\text{A}}(x^{(t-1)}))\text{,}
    \label{equ:PGD}
\end{equation}
where the increment of the loss is $\Delta\Loss_{\text{A}}=\Loss_{\text{A}}(x^{(t)})-\Loss_{\text{A}}(x^{(t-1)})=\eta\|\nabla_{x}\Loss_{\text{A}}(x^{(t-1)})\|_{q}$.

\subsection{The inner maximization problem in AT}
As shown in Table~\ref{table:6}, the adversarial objectives $\Loss_{\text{A}}$ are usually the CE loss between the adversarial prediction $f(x')$ and the target prediction $f(x)$ or $1_{y}$. Thus, to investigate the inner maximization problem of $\Loss_{\text{A}}$, we expand the gradient of CE loss w.r.t. $x'$ as below:
\begin{lemma}
\label{lemma2}
(The gradient of CE loss) Let $W_{ij}=W_{i}-W_{j}$ be the residual vector between two weights, and $z'=z(x';\bm{\omega})$ be the mapped feature of the adversarial example $x'$, then there is
\begin{equation}
    \nabla_{x'}\Loss_{\CE}(f(x'),f(x))=-\sum_{i\neq j}f(x)_{i}f(x')_{j}\nabla_{x'}(W_{ij}^{\top}z')\text{.}
    \label{eq:9}
\end{equation}
If $f(x)=1_{y}$ is the one-hot label vector, we have $\nabla_{x'}\Loss_{\CE}(f(x'),y)=-\sum_{l\neq y}f(x')_{l}\nabla_{x'}(W_{yl}^{\top}z')$.
\end{lemma}
Lemma~\ref{lemma2} indicates that the gradient of CE loss can be decomposed into the linear combination of the gradients on the residual logits $W_{ij}^{\top}z'$. Let $y^*$ be the predicted label on the finally crafted adversarial example $x^*$, where $y^{*}\neq y$. Based on the empirical observations~\citep{Goodfellow-et-al2016,pang2018towards}, we are justified to assume that $f(x)_{y}$ is much larger than $f(x)_{l}$ for $l\neq y$, and $f(x')_{y^*}$ is much larger than $f(x')_{l}$ for $l\not\in \{y,y^*\}$. Then we can approximate the linear combination in Eq.~(\ref{eq:9}) with the dominated term as
\begin{equation}
\nabla_{x'}\Loss_{\CE}(f(x'),f(x))\approx-f(x)_{y}f(x')_{y^*}\nabla_{x'}(W_{yy^*}^{\top}z')\text{, where }W_{yy^*}=W_{y}-W_{y^*}\text{.}
\label{equ:withoutHE}
\end{equation}
Let $\theta_{yy^*}'=\angle(W_{yy^*},z')$, there is $W_{yy^*}^{\top}z'=\|W_{yy^*}\|\|z'\|\cos(\theta_{yy^*}')$ and $W_{yy^*}$ does not depend on $x'$. Thus by substituting Eq.~(\ref{equ:withoutHE}) into Eq.~(\ref{equ:PGD}), the update direction of each attacking step becomes
\begin{equation}
\mathbb{U}_{p}[\nabla_{x'}\Loss_{\CE}(f(x'),f(x))]\approx-\mathbb{U}_{p}[\nabla_{x'}(\|z'\|\cos({\theta_{yy^*}'}))]\text{,}
\label{equ:8}
\end{equation}
where the factor $f(x)_{y}f(x')_{y^*}$ is eliminated according to the definition of $\mathbb{U}_{p}[\cdot]$. Note that Eq.~(\ref{equ:8}) also holds when $f(x)=1_{y}$, and the resulted adversarial objective is analogous to the C\&W attack~\citep{Carlini2016}.

\subsection{Benefits from feature normalization}
\label{BFN}
To investigate the effects of FN alone, we deactivate the WN operation in Eq.~(\ref{equ:4}) and denote
\begin{equation}
    \overline{f}(x)=\softmax({\mathbf{W}}^{\top}\widetilde{z})\text{.}
\end{equation}
Then similar to Eq.~(\ref{equ:8}), we can obtain the update direction of the attack with FN applied as
\begin{equation}
\mathbb{U}_{p}[\nabla_{x'}\Loss_{\CE}(\overline{f}(x'),\overline{f}(x))]\approx-\mathbb{U}_{p}[\nabla_{x'}(\cos({\theta_{yy^*}'}))]\text{.}
\label{equ:9}
\end{equation}

\begin{figure}[t]
\begin{center}
\vspace{-0.2cm}
\includegraphics[width=.9\columnwidth]{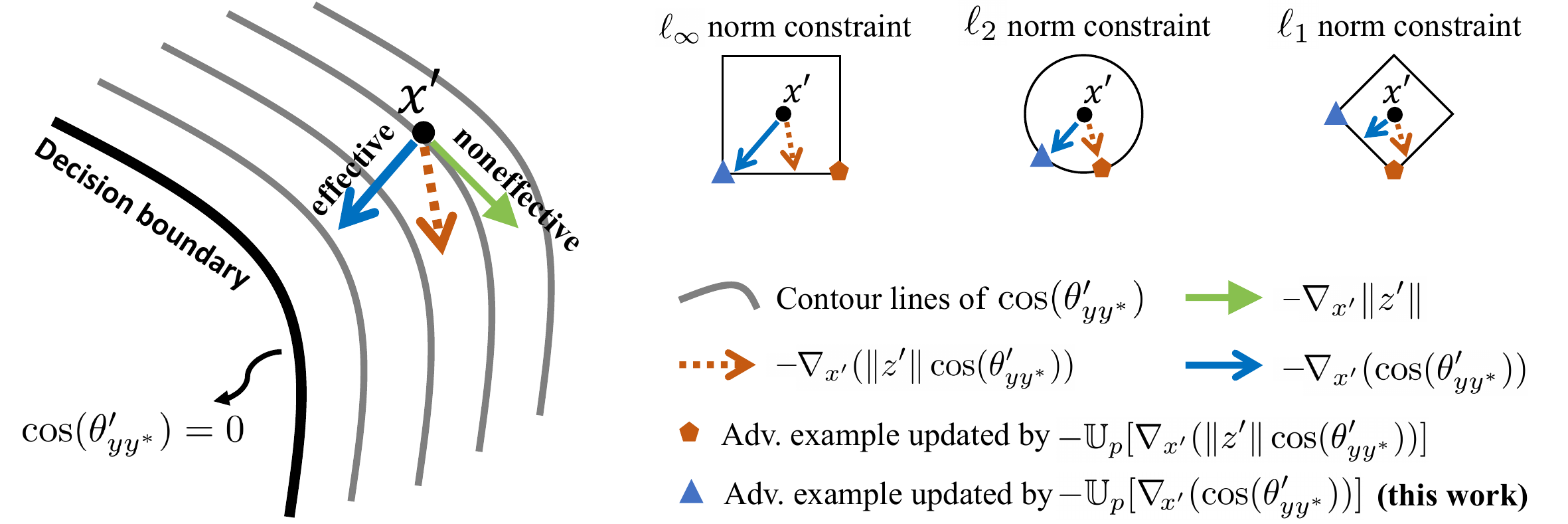}
\vspace{-0.1cm}
\caption{Intuitive illustration in the input space. When applying FN in $\Loss_{\text{A}}$, the adversary can take more effective update steps to move $x'$ across the decision boundary defined by $\cos(\theta_{yy^*}')=0$.}
\label{fig:1}
\end{center}
\vspace{-0.1cm}
\end{figure}

\textbf{More effective adversarial perturbations.} In the AT procedure, we prefer to craft adversarial examples more efficiently to reduce the computational burden~\citep{liu2020using,vivek2020single,wong2020fast}. As shown in Fig.~\ref{fig:1}, to successfully fool the model to classify $x'$ into the label $y^*$, the adversary needs to craft iterative perturbations to move $x'$ across the decision boundary defined by $\cos(\theta'_{yy^*})=0$. Under the first-order optimization, the most effective direction to reach the decision boundary is along $-\nabla_{x'}(\cos(\theta_{yy^*}'))$, namely, the direction with the steepest descent of $\cos(\theta'_{yy^*})$. In contrast, the direction of $-\nabla_{x'}\|z'\|$ is nearly tangent to the contours of $\cos(\theta'_{yy^*})$, especially in high-dimension spaces, which is noneffective as the adversarial perturbation. Actually from Eq.~(\ref{equ:1}) we can observe that when there is no bias term in the softmax layer, changing the norm $\|z'\|$ will not affect the predicted labels at all.

By comparing Eq.~(\ref{equ:8}) and Eq.~(\ref{equ:9}), we can find that applying FN in the adversarial objective $\Loss_{\text{A}}$ exactly removes the noneffective component caused by $\|z'\|$, and encourages the adversarial perturbations to be aligned with the effective direction $-\nabla_{x'}(\cos(\theta_{yy^*}'))$ under the $\ell_{p}$-norm constraint. This facilitate crafting adversarial examples with fewer iterations and improve the efficiency of the AT progress, as empirically verified in the left panel of Fig.~\ref{fig:3}. Besides, in Fig.~\ref{fig:1} we also provide three instantiations of the $\ell_{p}$-norm constraint. We can see that if we do not use FN, the impact of the noneffective component of $-\nabla_{x'}\|z'\|$ could be magnified under, e.g., the $\ell_{\infty}$-norm constraint, which could consequently require more iterative steps and degrade the training efficiency.

\textbf{Better learning on hard (adversarial) examples.} As to the benefits of applying FN in the training objective $\Loss_{\text{T}}$, we formally show that FN can promote learning on hard examples, as empirically observed in the previous work~\citep{ranjan2017l2}. In the adversarial setting, this property can promote the worst-case performance under potential adversarial threats. Specifically, the model parameters $\bm{\omega}$ is updated towards $-\nabla_{\bm{\omega}}\Loss_{\CE}$. When FN is not applied, we can use similar derivations as in Lemma~\ref{lemma2} to obtain
\begin{equation}\label{eq:No-FN}
    -\!\nabla_{\bm{\omega}}\Loss_{\CE}(f(x),y)=\sum_{l\neq y}f(x)_{l}\|W_{yl}\|(\cos({\theta_{yl}})\nabla_{\bm{\omega}}\|z\|\!+\!\|z\|\nabla_{\bm{\omega}}\cos({\theta_{yl}}))\text{,}
\end{equation}
where $W_{yl}=W_{y}-W_{l}$ and $\theta_{yl}=\angle(W_{yl},z)$. According to Eq.~(\ref{eq:No-FN}), when we use a mini-batch of data to update $\bm{\omega}$, the inputs with small $\nabla_{\bm{\omega}}\|z\|$ or $\nabla_{\bm{\omega}}\cos({\theta_{yl}})$ contribute less in the direction of model updating, which are qualitatively regarded as hard examples~\citep{ranjan2017l2}. This causes the training process to devote noneffective efforts to increasing $\|z\|$ for easy examples and consequently overlook the hard ones, which leads to vicious circles and could degrade the model robustness against strong adversarial attacks~\citep{rice2020overfitting,song2019improving}. As shown in Fig.~\ref{fig:2}, the hard examples in AT are usually the crafted adversarial examples, which are those we actually expect the model to focus on in the AT procedure. In comparison, when FN is applied, there is $\nabla_{\bm{\omega}}\|\widetilde{z}\|=0$, then $\bm{\omega}$ is updated towards
\begin{equation}
    \vspace{-0.cm}
    -\!\nabla_{\bm{\omega}}\Loss_{\CE}(\overline{f}(x),y)=\sum_{l\neq y} \overline{f}(x)_{l}\|W_{yl}\|\nabla_{\bm{\omega}}\cos({\theta_{yl}})\text{.}
    \label{equ:17}
    \vspace{-0.cm}
\end{equation}
In this case, due to the bounded value range $[-1,1]$ of cosine function, the easy examples will contribute less when they are well learned, i.e., have large $\cos(\theta_{yl})$, while the hard examples could later dominate the training. This causes a dynamic training procedure similar to curriculum learning~\citep{bengio2009curriculum}.

\vspace{-0.1cm}
\subsection{Benefits from weight normalization}
\vspace{-0.1cm}
In the AT procedure, we usually apply untargeted attacks~\citep{engstrom2018evaluating,madry2018towards}. Since we do not explicit assign targets, the resulted prediction labels and feature locations of the crafted adversarial examples will depend on the unbalanced semantic similarity among different classes. For example, the learned features of dogs and cats are usually closer than those between dogs and planes, so an untargeted adversary will prefer to fool the model to predict the cat label on a dog image, rather than the plane label~\citep{Moosavidezfooli2016}. To understand how the adversarial class biases affect training, assuming that we perform the gradient descent on a data batch $\mathcal{D}=\{(x^{k},y^{k})\}_{k\in[N]}$. Then we can derive that $\forall l\in[L]$, the softmax weight $W_{l}$ is updated towards
\begin{equation}
    -\nabla_{W_{l}}\Loss_{\CE}(\mathcal{D})=\sum_{x^k\in\mathcal{D}_{l}} z^{k}-\sum_{x^k\in\mathcal{D}}f(x^{k})_{l}\cdot z^{k}\text{,}
    \label{equ:18}
\end{equation}
where $\mathcal{D}_{l}$ is the subset of $\mathcal{D}$ with true label $l$. We can see that the weight $W_{l}$ will tend to have larger norm when there are more data or easy examples in class $l$, i.e., larger $\left|\mathcal{D}_{l}\right|$ or $\|z^{k}\|$ for $x^k\in\mathcal{D}_{l}$. Besides, if an input $x$ in the batch is adversarial, then $f(x)_{y}$ is usually small and consequently $z$ will have a large effect on the update of $W_{y}$. Since there is $W_{yy^*}^\top z<0$, $W_{y}$ will be updated towards $W_{y^*}$ both in norm and direction, which causes repetitive oscillation during training.

When applying WN, the update of $W_{l}$ will only depend on the averaged feature direction within each class, which alleviates the noneffective oscillation on the weight norm and speed up training~\citep{salimans2016weight}. Besides, when FN and WN are both applied, the inner products $\bm{W}^{\top}z$ in the softmax layer will become the angular metric $\cos\bm{\theta}$, as shown in Eq.~(\ref{equ:4}). Then we can naturally introduce AM to learn angularly more discriminative and robust features~\citep{elsayed2018large,yan2018deep}.

\subsection{Modifications to better utilize strong adversaries}
\label{modification}
In most of the AT procedures, the crafted adversarial examples will only be used once in a single training step to update the model parameters~\citep{madry2018towards}, which means hard adversarial examples may not have an chance to gradually dominate the training as introduced in Eq.~(\ref{equ:17}). Since $\nabla_{\bm{\omega}}\cos({\theta_{yy^*}})=-\sin({\theta_{yy^*}})\nabla_{\bm{\omega}}{\theta_{yy^*}}$, the weak adversarial examples around the decision boundary with $\theta_{yy^*}\sim 90^{\circ}$ have higher weights $\sin({\theta_{yy^*}})$. This makes the model tend to overlook the strong adversarial examples with large $\theta_{yy^*}$, which contain abundant information. To be better compatible with strong adversaries, an easy-to-implement way is to directly substitute $\softmax(\cos{\bm{\theta}})$ in Eq.~(\ref{equ:4}) with $\softmax(-\bm{\theta})$, using the $\arccos$ operator. We name this form of embedding as modified HE (\textbf{m-HE}), as evaluated in Table~\ref{table:19}.

\begin{table}[t]
\vspace{-0.cm}
  \caption{Classification accuracy (\%) on \textbf{CIFAR-10} under the \emph{white-box} threat model. The perturbation $\epsilon=0.031$, step size $\eta=0.003$. We highlight the best-performance model under each attack.}
  \vspace{-0.cm}
  \begin{center}
  \begin{small}
  \renewcommand*{\arraystretch}{1.2}
  \begin{tabular}{c|c|c|c|c|c|c|c|c|c}
  \hline
Defense&\! Clean \!& \!\!PGD-20\! \!&\!\!PGD-500\!\! &\! \!MIM-20\!\! &\! \!FGSM \!\!& \!DeepFool\! & \!C\&W\! & \!FeaAtt.\! & \!FAB\! \\
\hline
PGD-AT & {86.75} & 53.97 & 51.63 & 55.08 & 59.70 & 57.26 & 84.00 & 52.38 & 51.23\\
\!\!\!\!PGD-AT+\textbf{HE}\!\!\!\! & 86.19 & {59.36} & {57.59} & {60.19} & {63.77} & {61.56} & {84.07} & {52.88} & \textbf{54.45}\\
\hline
ALP & 87.18 & 52.29 & 50.13 & 53.35 & 58.99 & 59.40 & 84.96 & {49.55} & 50.54\\
\!\!\!\!ALP+\textbf{HE}\!\!\!\! & \textbf{89.91} & {57.69} & {51.78} & {58.63} & {65.08} & \textbf{65.19} & \textbf{87.86} &48.64 & {51.86}\\
\hline
TRADES & 84.62 & 56.48 & 54.84 & 57.14 & 61.02 & {60.70} & 81.13 & 55.09 & {53.58}\\
\!\!\!\!TRADES+\textbf{HE} \!\!\!\!& {84.88} & \textbf{62.02} & \textbf{60.75} & \textbf{62.71} & \textbf{65.69} & 60.48 & {81.44} & \textbf{58.13} & 53.50\\
  \hline
   \end{tabular}
  \end{small}
  \end{center}
  \label{table:7}
  \vspace{-.0cm}
\end{table}

\begin{figure}[t]
\vspace{-0.in}
\begin{minipage}[t]{.48\linewidth}
\captionof{table}{Validation of combining FastAT and FreeAT with HE and m-HE on \textbf{CIFAR-10}. We report the accuracy (\%) on clean and PGD, as well as the total training time (min).}
\vspace{-0.1cm}
  \begin{center}
  \begin{small}
  \renewcommand*{\arraystretch}{1.2}
  \begin{tabular}{c|c|c|c|c}
  \hline
\!Defense&\! \!Epo.\!\! & \!Clean\! &\!\!PGD-50\!\!  & \!\!Time\!\!\\
\hline
\!FastAT& {30} & \textbf{83.80} & 46.40 &  \!\!\emph{11.38} \!\!\\
\!FastAT \!+\! \textbf{HE}& {30} & 82.58 & {52.55} &  \!\!\emph{11.48} \!\!\\
\!\!\!FastAT \!+ \!\textbf{m-HE}\!\!& {30} & 83.14 & \textbf{53.49} &  \!\!\emph{11.49} \!\!\\
\hline
\!FreeAT& {10} & {77.21} & 46.14 &  \!\!\emph{15.78}  \!\!\\
\!FreeAT \!+\! \textbf{HE}& {10} & 76.85 & {50.98} &  \!\!\emph{15.87} \!\!\\
\!\!\!FreeAT \!+ \!\textbf{m-HE}\!\!& {10} & \textbf{77.59} & \textbf{51.85} &  \!\!\emph{15.91} \!\!\\
\hline
   \end{tabular}
  \end{small}
  \end{center}
  \label{table:19}
  \end{minipage}
  \hspace{0.4cm}
\begin{minipage}[t]{.48\linewidth}
\captionof{table}{{Top-1} classification accuracy (\%) on \textbf{ImageNet} under the \emph{white-box} threat model.}
\vspace{-0.2cm}
  \begin{center}
  \begin{small}
  \renewcommand*{\arraystretch}{1.2}
  \begin{tabular}{c|c|c|c|c}
  \hline
Model& Method &\!\!\!Clean\!\! \!&\!\!\!PGD-10\!\!\! & \!\!\!PGD-50\!\!\!\\
\hline
\multirow{2}{*}{\!\!\!ResNet-50\!\!\!}&FreeAT& \!\!\!60.28\!\!\! & 32.13 &31.39\\
&\!\!FreeAT \!+ \!\textbf{HE}\!\!&\!\!\!\textbf{61.83}\!\!\! & \textbf{40.22} & \textbf{39.85}\\
\hline

\multirow{2}{*}{\!\!\!ResNet-152\!\!\!}&FreeAT& \!\!\!65.20\!\!\! & 36.97 & 35.87\\
&\!\!FreeAT \!+ \!\textbf{HE}\!\!&\!\!\!\textbf{65.41}\!\!\!&\textbf{43.24}&\textbf{42.60}\\
\hline
\multirow{2}{*}{\!\!\!WRN-50-2\!\!\!}&FreeAT &\!\!\! 64.18 \!\!\!&36.24 & 35.38\\
&\!\!FreeAT \!+\! \textbf{HE}\!\! &\!\!\!\textbf{65.28}\!\!\! &\textbf{43.83} &\textbf{43.47} \\
\hline
\multirow{2}{*}{\!\!\!\!WRN-101-2\!\!\!}&FreeAT& \!\!\!66.15 \!\!\!& 39.35 & 38.23 \\
&\!\!FreeAT \!+\! \textbf{HE}\!\!&\!\!\! \textbf{66.37} \!\!\!& \textbf{45.35} & \textbf{45.04} \\
\hline
  \end{tabular}
  \end{small}
  \end{center}
  \label{table:9}
\end{minipage}
  \vspace{-0.cm}
  \end{figure}

\section{Experiments}
\label{Experiements}

\textbf{CIFAR-10~\citep{Krizhevsky2012} setup.} We apply the wide residual network WRN-34-10 as the model architecture~\cite{zagoruyko2016wide}. For each AT framework, we set the maximal perturbation $\epsilon=8/255$, the perturbation step size $\eta=2/255$, and the number of iterations $K=10$. We apply the momentum SGD~\citep{qian1999momentum} optimizer with the initial learning rate of $0.1$, and train for 100 epochs. The learning rate decays with a factor of $0.1$ at $75$ and $90$ epochs, respectively. The mini-batch size is $128$. 
Besides, we set the regularization parameter $1/\lambda$ as $6$ for TRADES, and set the adversarial logit pairing weight as 0.5 for ALP~\citep{kannan2018adversarial,zhang2019theoretically}. The scale $s=15$ and the margin $m=0.2$ in HE, where different $s$ and $m$ correspond to different trade-offs between the accuracy and robustness, as detailed in Appendix {\color{red} C.3}.

\textbf{ImageNet~\citep{deng2009imagenet} setup.} We apply the framework of free adversarial training (\textbf{FreeAT}) in \citet{shafahi2019adversarial}, which has a similar training objective as PGD-AT and can train a robust model using four GPU workers. We set the repeat times $m=4$ in FreeAT. The perturbation $\epsilon=4/255$ with the step size $\eta=1/255$. We train the model for 90 epochs with the initial learning rate of $0.1$, and the mini-batch size is $256$. The scale $s=10$ and the margin $m=0.2$ in HE.\footnote{Code is available at \url{https://github.com/ShawnXYang/AT_HE}.}


%

\vspace{-0.1cm}
\subsection{Performance under white-box attacks}
\vspace{-0.1cm}

On the CIFAR-10 dataset, we test the defenses under different attacks including FGSM~\citep{Goodfellow2014}, PGD~\citep{madry2018towards}, MIM~\citep{Dong2017}, Deepfool~\citep{Moosavidezfooli2016}, C\&W ($l_\infty$ version)~\citep{Carlini2016}, feature attack~\citep{featureattack}, and FAB~\citep{croce2019minimally}. We report the classification accuracy in Table~\ref{table:7} following the evaluation settings in~\citet{zhang2019theoretically}. We denote the iteration steps behind the attacking method, e.g., 10-step PGD as PGD-10. To verify that our strategy is generally compatible with previous work on accelerating AT, we combine HE with the one-step based FreeAT and fast adversarial training (\textbf{FastAT}) frameworks~\citep{wong2020fast}.
We provide the accuracy and training time results in Table~\ref{table:19}. We can see that the operations in HE increase negligible computation, even in the cases pursuing extremely fast training. Besides, we also evaluate embedding \textbf{m-HE} (introduced in Sec.~\ref{modification}) and find it more effective than HE when combining with PGD-AT, FreeAT and Fast AT that exclusively train on adversarial examples. On the ImageNet dataset, we follow the evaluation settings in~\citet{shafahi2019adversarial} to test under PGD-10 and PGD-50, as shown in Table~\ref{table:9}.

\begin{table*}[t]
\footnotesize
\setlength{\tabcolsep}{3pt}
  \vspace{-0.5cm}
  \caption{Top-1 classification accuracy (\%) on \textbf{CIFAR-10-C} and \textbf{ImageNet-C}. The models are trained on the original datasets CIFAR-10 and ImageNet, respectively. Here 'mCA' refers to the mean accuracy averaged on different corruptions and severity. Full version of the table is in Appendix {\color{red} C.6}.}
  \vspace{-0.2cm}
  \begin{center}
  \begin{tabular}{c|c|cccc|cccc|cccc}
  \hline
\multirow{2}{*}{Defense}& \multirow{2}{*}{mCA} & \multicolumn{4}{c|}{\emph{Blur}}& \multicolumn{4}{c|}{\emph{Weather}}&\multicolumn{4}{c}{\emph{Digital}}\\ 
&  & \!Defocus\!\!& Glass& \!\!Motion\!\!& \!Zoom\! &Snow & Frost & Fog & \!Bright\! &\!Contra\! & \!Elastic\!& Pixel& \!JPEG\!\\
\hline
\multicolumn{14}{c}{\textbf{CIFAR-10-C}}\\  
\hline
\!\!PGD-AT\! & 77.23   & 81.84 & 79.69 & 77.62 & 80.88 & 81.32 & 77.95 & 61.70 & 84.05 & 44.55 & 80.79 & 84.76 & 84.35\\
\!\!PGD-AT \!+\! \textbf{HE}\! & \textbf{77.29}  & 81.86 & 79.45 & 78.17 & 80.87 & 80.77 & 77.98 & 62.45 & 83.67 & 45.11 & 80.69 & 84.16 & 84.10\\
\hline
\!\!ALP\! & 77.73 &  81.94 & 80.31 & 78.23 & 80.97 & 81.74 & 79.26 & 61.51 & 84.88 & 45.86 & 80.91 & 85.09& 84.68\\
\!\!ALP\! + \!\textbf{HE}\! & \textbf{80.55} &  80.87& 85.23 & 81.26 & 84.43 & 85.14 & 83.89 & 68.83 & 88.33 & 50.74 & 84.44 & 87.44 & 87.28\\
\hline
\!\!TRADES\! & 75.36 &  79.84 & 77.72 & 76.34 & 78.66 & 79.52 & 76.94 & 59.68 & 82.06 & 43.80 & 78.53 & 82.65& 82.31\\
\!\!TRADES \!+ \!\textbf{HE}\! & \textbf{75.78} & 80.55 & 77.61 & 77.26 & 79.62 & 79.23 & 76.53 & 61.39 & 82.33 & 45.04 & 79.29 & 82.50& 82.40\\
  \hline
\multicolumn{14}{c}{\textbf{ImageNet-C}}\\  
\hline
\!\!FreeAT\! & 28.22 & 19.15& 26.63& 25.75& 28.25& 23.03& 23.47& 3.71& 45.18& 5.40& 41.76& 48.78& 52.55\\

\!\!FreeAT \!+ \!\textbf{HE}\! &\textbf{30.04} & 21.16& 29.28& 28.08 &30.76& 26.62& 28.35& 5.34 & 49.88& 7.03& 44.72& 51.17& 55.05\\

  \hline
   \end{tabular}
  \end{center}
  \label{table:7-2}
  \vspace{-.cm}
\end{table*}

\begin{figure}[t]
\begin{center}
\vspace{-0.4cm}
\includegraphics[width=1.00\columnwidth]{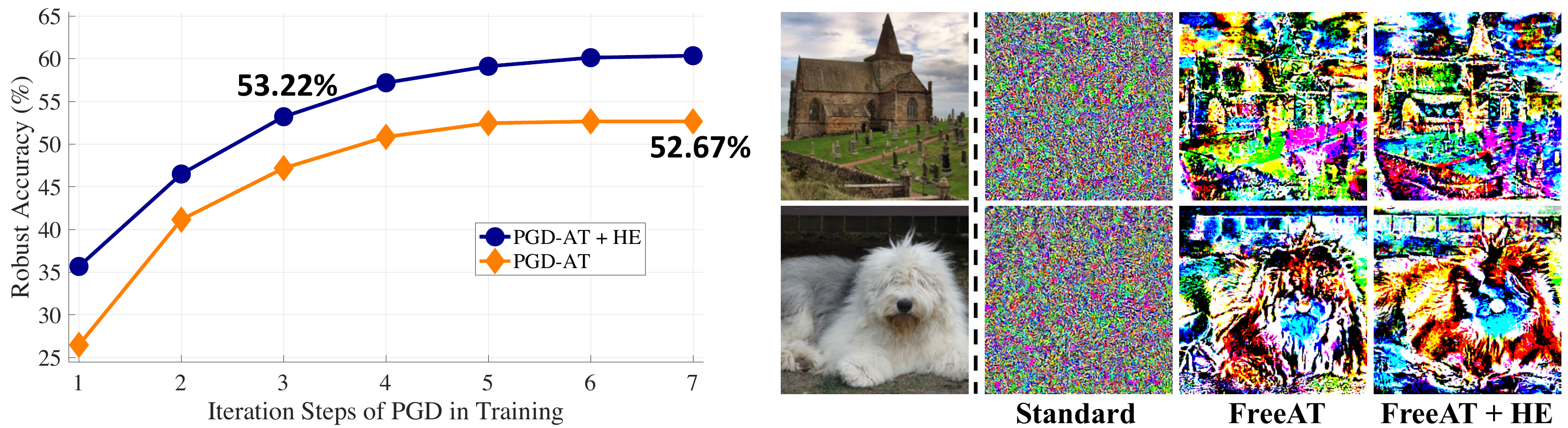}
\vspace{-0.4cm}
\caption{\emph{Left.} Accuracy (\%) under PGD-20, where the models are trained by PGD-AT with different iteration steps on \textbf{CIFAR-10}. \emph{Right.} Visualization of the adversarial perturbations on \textbf{ImageNet}.} 
\label{fig:3}
\end{center}
\vspace{-0.25cm}
\end{figure}

\textbf{Ablation studies.} To investigate the individual effects caused by the three components FN, WN, and AM in HE, we perform the ablation studies for PGD-AT on CIFAR-10, and attack the trained models with PGD-20. We get the clean and robust accuracy of 86.43\% / 54.46\% for PGD-AT+FN, and 87.28\% / 53.93\% for PGD-AT+WN. In contrast, when we apply both the FN and WN, we can get the result of 86.20\% / 58.95\%. For TRADES, applying appropriate AM can increase $\sim$2\% robust accuracy compared to only applying FN and WN. Detailed results are included in the Appendix {\color{red} C.3}.

\textbf{Adaptive attacks.} A generic PGD attack apply the cross-entropy loss on the model prediction as the adversarial objective, as reported in Table~\ref{table:7}. To exclude the potential effect of gradient obfuscation~\citep{athalye2018obfuscated}, we construct an adaptive version of the PGD attack against our methods, which uses the training loss in Eq.~(\ref{CE_HE}) with the scalar $s$ and margin $m$ as the adversarial objective. In this case, when we apply the adaptive PGD-20 / PGD-500 attack, we will get a accuracy of $55.25\%$ / $52.54\%$ for PGD-AT+HE, which is still higher than the accuracy of $53.97\%$ / $51.63\%$ for PGD-AT (quote from Table~\ref{table:7}).

\begin{figure}[t]
\vspace{-0.cm}
\begin{minipage}[t]{.57\linewidth}
\captionof{table}{Classification accuracy (\%) on the clean test data, and under two benchmark attacks RayS and AutoAttack.}
\vspace{-0.cm}
  \begin{center}
  \begin{small}
  \renewcommand*{\arraystretch}{1.3}
  \begin{tabular}{c|c|c|c|c}
  \hline
  {Method} & Architecture & Clean & RayS & AA \\
   \hline
   \multirow{2}{*}{PGD-AT+\textbf{HE}} & \!\!\! WRN-34-10 \!\!\! & 86.25 & 57.8 & 53.16 \\
   & \!\!\! WRN-34-20 \!\!\! & 85.14 & 59.0 & 53.74\\
    \hline
      \end{tabular}
  \label{tab:benchmark}%
  \end{small}
  \end{center}
  \end{minipage}
  \hspace{0.4cm}
\begin{minipage}[t]{.37\linewidth}
\captionof{table}{Attacking standardly trained WRN-34-10 with or without FN.}
\vspace{-0.2cm}
  \begin{center}
  \begin{small}
  \begin{tabular}{c|c|c}
  \hline
   Attack  &  FN  &  Acc. (\%)  \\
   \hline
    \multirow{2}{*}{PGD-1}  &  \ding{54}  & \!\! 67.09 \!\! \\
   &  \ding{52}   & \!\! \textbf{62.89} \!\! \\
   \hline
    \multirow{2}{*}{PGD-2}  &  \ding{54}  & \!\! 50.37 \!\!  \\
   &  \ding{52}   & \textbf{33.75} \!\! \\
    \hline
      \end{tabular}
  \label{tab:PGD12}%
  \end{small}
  \end{center}
\end{minipage}
  \vspace{-0.cm}
  \end{figure}

\textbf{Benchmark attacks.} We evaluate our enhanced models under two stronger benchmark attacks including $\textrm{RayS}$~\citep{chen2020rays} and $\textrm{AutoAttack}$~\citep{croce2020reliable} on CIFAR-10. We train WRN models via PGD-AT+HE, with weight decay of $5\times10^{-4}$~\citep{pang2020bag}. For RayS, we evaluate on $1,000$ test samples due to the high computation. The results are shown in Table~\ref{tab:benchmark}, where the trained WRN-34-20 model achieves the state-of-the-art performance (no additional data) according to the reported benchmarks.


\subsection{Performance under black-box attacks}

\textbf{Query-based Black-box Attacks.} ZOO~\citep{chen2017zoo} proposes to estimate the gradient at each coordinate $\bm{e}_i$ as $\hat{g}_i$, with a small finite-difference $\sigma$. In experiments, we randomly select one sampled coordinate to perform one update with $\hat{g}_i$, and adopt the C\&W optimization mechanism based on the estimated gradient. We set $\sigma$ as $10^{-4}$ and max queries as $20,000$. SPSA~\citep{uesato2018adversarial} and NES~\citep{ilyas2018black} can make a full gradient evaluation by drawing random samples and obtaining the corresponding loss values. NES randomly samples from a Gaussian distribution to acquire the direction vectors while SPSA samples from a Rademacher distribution. In experiments, we set the number of random samples $q$ as 128 for every iteration and $\sigma = 0.001$. We show the robustness of different iterations against untargeted score-based ZOO, SPSA, and NES in Table~\ref{table:7-1}, where details on these attacks are in Appendix {\color{red} B.2}.

We include detailed experiments on the \textbf{transfer-based black-box attacks} in Appendix {\color{red} C.4}. As expected, these results show that embedding HE can generally provide promotion under the black-box threat models~\citep{carlini2019evaluating}, including the transfer-based and the query-based black-box attacks.

\begin{table*}[t]
\vspace{-0.1cm}
  \caption{Classification accuracy (\%) under different \emph{black-box} query-based attacks on \textbf{CIFAR-10}.}
  \vspace{-0.3cm}
  \begin{center}
  \begin{small}
  \begin{tabular}{c|c|c|c|c|c|c|c}
  \hline
Method & Iterations & PGD-AT & PGD-AT + \textbf{HE} & ALP & ALP + \textbf{HE} & TRADES & TRADES + \textbf{HE} \\
\hline
ZOO & - & 73.47 & 74.65 & 72.70 & 74.90 & 71.92 & 74.65 \\
\hline
\multirow{3}{*}{SPSA} & \emph{20} & 73.64 & 73.66 & 73.11 & 74.45 & 72.12 & 72.36\\
& \emph{50} & 68.93 & 69.31 & 68.39 & 68.28 & 68.20 & 68.42 \\
& \emph{80} & 65.67 & 65.97 & 65.14 & 63.89 & 65.32 & 65.62\\
\hline
\multirow{3}{*}{NES} & \emph{20} & 74.68 & 74.87 &74.41 & 75.91 & 73.40 & 73.47 \\
& \emph{50} & 71.22 & 71.48 & 70.81 & 71.11 & 70.29 & 70.53 \\
& \emph{80} & 69.16 & 69.92 & 68.88 & 68.53 & 68.83 & 69.06 \\
\hline

   \end{tabular}
  \end{small}
  \end{center}
  \label{table:7-1}
  \vspace{-0cm}
\end{table*}

\begin{figure}[t]
\begin{center}
\vspace{-0.5cm}
\includegraphics[width=1.00\columnwidth]{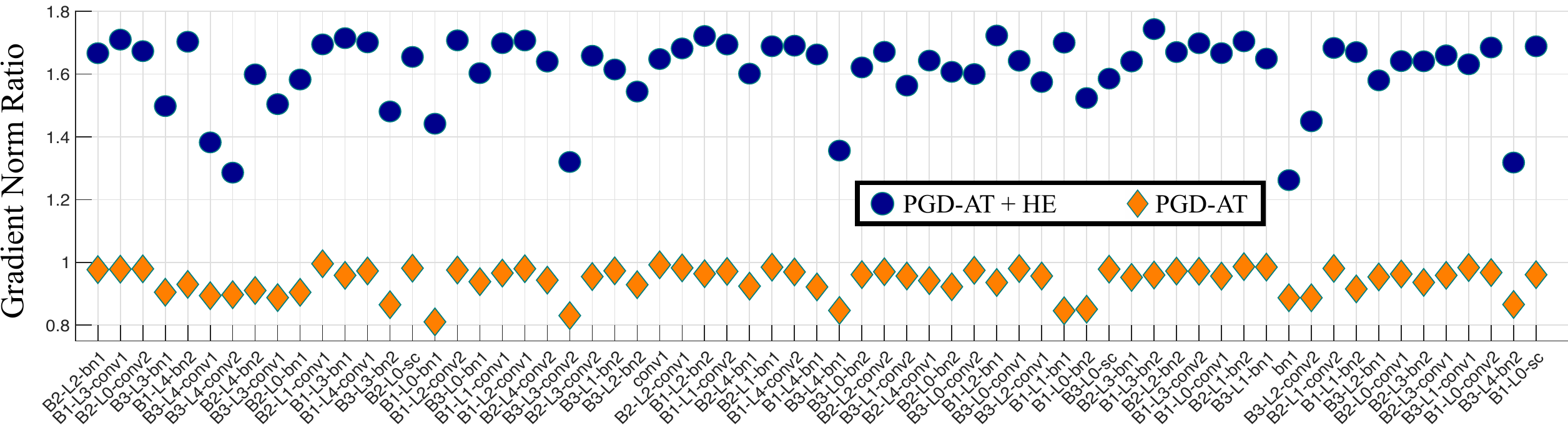}
\vspace{-0.5cm}
\caption{The ratios of $\mathbb{E}(\|\nabla_{\bm{\omega}}\Loss(x^*)\|/\|\nabla_{\bm{\omega}}\Loss(x)\|)$ w.r.t. different parameters $\omega$, where $x^*$ and $x$ are the adversarial example and its clean counterpart. Higher ratio values indicate more attention on the adversarial examples. 'B' refers to block, 'L' refers to layer, 'conv' refers to convolution.}
\label{fig:2}
\end{center}
\vspace{-0.cm}
\end{figure}

\subsection{Performance under general-purpose attacks}
It has been shown that the adversarially trained models could be vulnerable to rotations~\citep{engstrom2017rotation}, image corruptions~\citep{ford2019adversarial} or affine attacks~\citep{tramer2019adversarial}. Therefore, we evaluate on the benchmarks with distributional shifts: CIFAR-10-C and ImageNet-C~\citep{hendrycks2019benchmarking}. As shown in Table~\ref{table:7-2}, we report the classification accuracy under each corruption averaged on five levels of severity, where the models are WRN-34-10 trained on CIFAR-10 and ResNet-50 trained on ImageNet, respectively. Here we adopt accuracy as the metric to be consistent with other results, while the reported values can easily convert into the corruption error metric~\citep{hendrycks2019benchmarking}. We can find that our methods lead to better robustness under a wide range of corruptions that are not seen in training, which prevents the models from overfitting to certain attacking patterns.

\subsection{More empirical analyses}

As shown in the left panel of Fig.~\ref{fig:3}, we separately use PGD-1 to PGD-7 to generate adversarial examples in training, then we evaluate the trained models under the PGD-20 attack. Our method requires fewer iterations of PGD to achieve certain robust accuracy, e.g., applying 3-step PGD in PGD-AT+HE is more robust than applying 7-step PGD in PGD-AT, which largely reduces the necessary computation~\citep{wang2019convergence}. To verify the mechanism in Fig.~\ref{fig:1}, we attack a standardly trained WRN-34-10 (no FN applied) model, applying PGD-1 and PGD-2 with or without FN in the adversarial objective. We report the accuracy in Table~\ref{tab:PGD12}. As seen, the attacks are more efficient with FN, which suggest that the perturbations are crafted along more effective directions.

Besides, previous studies observe that the adversarial examples against robust models exhibit salient data characteristics~\citep{ilyas2019adversarial,santurkar2019image,tao2018attacks,tsipras2018robustness, zhang2019interpreting}. So we visualize the untargeted perturbations on ImageNet, as shown in the right panel of Fig.~\ref{fig:3}. We can observe that the adversarial perturbations produced for our method have sharper profiles and more concrete details, which are better aligned with human perception. Finally in Fig.~\ref{fig:2}, we calculate the norm ratios of the loss gradient on the adversarial example to it on the clean example. The model is trained for $70$ epochs on CIFAR-10 using PGD-AT. The results verify that our method can prompt the training procedure to assign larger gradients on the crafted adversarial examples, which would benefit robust learning.

\vspace{-0.26cm}
\section{Conclusion}
\vspace{-0.2cm}
In this paper, we propose to embed the HE mechanism into AT, in order to enhance the robustness of the adversarially trained models. We analyze the intriguing benefits induced by the interaction between AT and HE from several aspects. It is worth clarifying that empirically our HE module has varying degrees of adaptability on combining different AT frameworks, depending on the specific training principles. Still, incorporating the HE mechanism is generally conducive to robust learning and compatible with previous strategies, with little extra computation and simple code implementation.




\section*{Broader Impact}
When deploying machine learning methods into the practical systems, the adversarial vulnerability can cause a potential security risk, as well as the negative impact on the crisis of confidence by the public. To this end, this inherent defect raises the requirements for reliable, general, and lightweight strategies to enhance the model robustness against malicious, especially adversarial attacks. In this work, we provide a simple and efficient way to boost the robustness of the adversarially trained models, which contributes to the modules of constructing more reliable systems in different tasks.

\section*{Acknowledgements}
This work was supported by the National Key Research and Development Program of China (No.2020AAA0104304), NSFC Projects (Nos. 61620106010, 62076147, U19B2034, U1811461), Beijing
Academy of Artificial Intelligence (BAAI), Tsinghua-Huawei Joint Research Program, a grant from Tsinghua Institute for Guo Qiang, Tiangong Institute for Intelligent Computing, and the NVIDIA NVAIL Program with GPU/DGX Acceleration.

\bibliography{main}
\bibliographystyle{plainnat}

\clearpage
\appendix
\section{Proofs}

\subsection{Proof of Lemma 1}
\begin{lemmaappendix}
Given a loss function $\Loss$ and under the first-order Taylor expansion, the solution of
\begin{equation*}
    \max_{\|x'-x\|_{p}\leq \epsilon}\Loss(x')
\end{equation*}
is $x^*=x+\epsilon\mathbb{U}_{p}(\nabla\Loss(x))$. Furthermore, there is $\Loss(x^*)=\Loss(x)+\epsilon\|\nabla\Loss(x)\|_{q}$, where $\|\cdot\|_{q}$ is the dual norm of $\|\cdot\|_{p}$. 
\end{lemmaappendix}

\emph{Proof.} We denote $x'=x+\epsilon v$, where $\|v\|_{p}\leq 1$. Then we know that $\|x'-x\|_{p}\leq \epsilon$. Under the first-order Taylor expansion, there is
\begin{equation*}
\begin{split}
    \max_{\|x'-x\|_{p}\leq \epsilon}\Loss(x')&=\max_{\|v\|_{p}\leq 1}\left[\Loss(x)+\epsilon v^{\top}\nabla\Loss(x)\right]\\
    &=\Loss(x)+\epsilon \max_{\|v\|_{p}\leq 1} v^{\top}\nabla\Loss(x)\text{.}
\end{split}
\end{equation*}
According to the definition of the dual norm~\citep{boyd2004convex}, there is $\max_{\|v\|_{p}\leq 1} v^{\top}\nabla\Loss(x)=\|\nabla\Loss(x)\|_{q}$, where where $\|\cdot\|_{q}$ is the dual norm of $\|\cdot\|_{p}$. Thus we prove that $\Loss(x^*)=\Loss(x)+\epsilon\|\nabla\Loss(x)\|_{q}$ and $x^*=x+\epsilon\mathbb{U}_{p}(\nabla\Loss(x))$.
\qed

\subsection{Proof of Lemma 2}

\begin{lemmaappendix}
By derivations, there is
\begin{equation*}
    \nabla_{x'}\Loss_{\CE}(f(x'),f(x))=-\sum_{i\neq j}f(x)_{i}f(x')_{j}\nabla_{x'}(W_{ij}^{\top}z')\text{,}
\end{equation*}
where $W_{ij}=W_{i}-W_{j}$, $z'=z(x';\bm{\omega})$. When $f(x)=1_{y}$, we have $\nabla_{x'}\Loss_{\CE}(f(x'),y)=-\sum_{l\neq y}f(x')_{l}\nabla_{x'}(W_{yl}^{\top}z')$.
\end{lemmaappendix}

\emph{Proof.} By derivations, there is
\begin{equation*}
\begin{split}
    &-\!\nabla_{x'}\Loss_{\CE}(f(x'),f(x))\\
    &\quad=\nabla_{x'}\left(f(x)^{\top}\log f(x')\right)\\
    &\quad=\sum_{i\in[L]}f(x)_{i}\nabla_{x'}\log(f(x')_{i})\\
    &\quad=\sum_{i\in[L]}f(x)_{i}\nabla_{x'}\log\left(\frac{\exp(W_{i}^{\top}z')}{\sum_{j\in[L]}\exp(W_{j}^{\top}z')}\right)\\
    &\quad=\sum_{i\in[L]}f(x)_{i}\nabla_{x'}\left({W_{i}^{\top}z'}-\log\left({\sum_{j\in[L]}\exp(W_{j}^{\top}z')}\right)\right)\\
    &\quad=\sum_{i\in[L]}f(x)_{i}\left(\nabla_{x'}{(W_{i}^{\top}z')}-{\sum_{j\in[L]}f(x')_{j}\nabla_{x'}(W_{j}^{\top}z')}\right)\\
    &\quad=\sum_{i\in[L]}f(x)_{i}\left({\sum_{j\neq i}f(x')_{j}\nabla_{x'}(W_{ij}^{\top}z')}\right)\\
    &\quad=\sum_{i\neq j}f(x)_{i}f(x')_{j}\nabla_{x'}(W_{ij}^{\top}z')\text{.}
\end{split}
\end{equation*}
Specially, when $f(x)=1_{y}$, we can obtain based on the above formulas that
\begin{equation*}
    \nabla_{x'}\Loss_{\CE}(f(x'),y)=-\sum_{l\neq y}f(x')_{l}\nabla_{x'}(W_{yl}^{\top}z')\text{.}
\end{equation*}
\qed

\vspace{-0.3cm}
\section{Related work}
In this section, we extensively introduce the related work in the adversarial setting, including the adversarial threat models (Sec.~\ref{b1}), the adversarial attacks (Sec.~\ref{b2}), the adversarial training strategy (Sec.~\ref{b3}), and some recent work on combining metric learning with adversarial training (Sec.~\ref{b4}).

\subsection{Adversarial threat models}
\label{b1}
Now we introduce different threat models in the adversarial setting following the suggestions in \citet{carlini2019evaluating}. Specifically, a threat model includes a set of assumptions about the adversary's goals, capabilities, and knowledge.

Adversary's goals could be simply fooling the classifiers to misclassify, which is referred to as \emph{untargeted mode}. On the other hand, the goals can be more aggressive to make the model misclassify from a source class into
a target class, which is referred to as \emph{targeted mode}.

Adversary's capabilities describe the constraints imposed on the attackers. For the $\ell_{p}$ bounded threat models, adversarial examples require the perturbation $\delta$ to be bounded by a preset threshold $\epsilon$ under $\ell_p$-norm, i.e., $\|\delta\|_{p}\leq\epsilon$.

Adversary's knowledge tells what knowledge the adversary is assumed to own. Typically, there are four settings when evaluating a defense method:
\begin{itemize}
\item \emph{Oblivious adversaries} are not aware of the existence of the defense $D$ and generate adversarial examples based on the unsecured classification model $F$~\citep{carlini2017adversarial}.
\item \emph{White-box adversaries} know the scheme and parameters of $D$, and can design adaptive methods to attack both the model $F$ and the defense $D$ simultaneously~\citep{athalye2018obfuscated}.
\item \emph{Black-box adversaries} have no access to the parameters of the defense $D$ or the model $F$ with varying degrees of black-box access~\citep{Dong2017}.
\item \emph{General-purpose adversaries} apply general transformations or corruptions on the images, which are related to traditional research topics on the input invariances~\citep{hendrycks2019using,zhang2019making}.
\end{itemize}

\subsection{Adversarial attacks}
\label{b2}
Below we show the details of the attack methods that we test on in our experiments. For clarity, we only introduce the untargeted attacks. The descriptions below mainly adopt from~\citet{dong2019benchmarking}. 

\textbf{FGSM} \citep{Goodfellow2014} generates an untargeted adversarial example under the $\ell_{\infty}$ norm as 
\begin{equation}
    \bm{x}^{adv} = \bm{x} + \epsilon\cdot\mathrm{sign}(\nabla_{\bm{x}}\mathcal{L}_{\text{CE}}(\bm{x},y))\text{.}
\end{equation}

\textbf{BIM} \citep{kurakin2016} extends FGSM by iteratively taking multiple small gradient updates as
\begin{equation}
\label{eq:bim}
    \bm{x}_{t+1}^{adv} = \mathrm{clip}_{\bm{x},\epsilon} \big(\bm{x}_t^{adv} + \eta\cdot\mathrm{sign}(\nabla_{\bm{x}}\mathcal{L}_{\text{CE}}(\bm{x}_t^{adv},y))\big),
\end{equation}
where $\mathrm{clip}_{\bm{x},\epsilon}$ projects the adversarial example to satisfy the $\ell_{\infty}$ constrain and $\eta$ is the step size.

\textbf{PGD} \citep{madry2018towards} is similar to BIM except that the initial point $\bm{x}_0^{adv}$ is uniformly sampled from the neighborhood around the clean input $x$, which can cover wider diversity of the adversarial space~\citep{wong2020fast}.

\textbf{MIM} \citep{Dong2017} integrates a momentum term into BIM with the decay factor $\mu=1.0$ as 
\begin{equation}
    \bm{g}_{t+1} = \mu \cdot \bm{g}_t + \frac{\nabla_{\bm{x}}\mathcal{L}_{\text{CE}}(\bm{x}_t^{adv},y)}{\|\nabla_{\bm{x}}\mathcal{L}_{\text{CE}}(\bm{x}_t^{adv},y)\|_1}\text{,}
\end{equation}
where the adversarial examples are updated by
\begin{equation}
    \bm{x}^{adv}_{t+1}=\mathrm{clip}_{\bm{x},\epsilon}(\bm{x}^{adv}_t+\alpha\cdot\mathrm{sign}(\bm{g}_{t+1}))\text{.}
\end{equation}
MIM has good performance as a transfer-based attack, which won the NeurIPS 2017 Adversarial Competition~\citep{kurakin2018competation}. We set the step size $\eta$ and the number of iterations identical to those in BIM.

\textbf{DeepFool} \citep{Moosavidezfooli2016} is also an iterative attack method, which generates an adversarial example on the decision boundary of a classifier with the minimum perturbation. We set the maximum number of iterations as $100$ in DeepFool, and it will early stop when the solution at an intermediate iteration is already adversarial.

\textbf{C\&W} \citep{Carlini2016} is a powerful optimization-based attack method, which generates an $\ell_2$ adversarial example $\bm{x}^{adv}$ by solving
\begin{equation}
\label{eq:cw}
\begin{split}
     \argmin_{\bm{x}'} \big\{c\cdot\max(Z(\bm{x}')_y &- \max_{i\neq y}Z(\bm{x}')_i, 0) \\
     & + \|\bm{x}'-\bm{x}\|_2^2 \big\}\text{,}
\end{split}
\end{equation}
where $Z(\bm{x}')$ is the logit output of the classifier and $c$ is a constant. This optimization problem is solved by an Adam \citep{Kingma2014} optimizer. $c$ is found by binary search. The C\&W attack can also be applied under the $\ell_{\infty}$ threat model with the adversarial loss function $\max(Z(\bm{x}')_y - \max_{i\neq y}Z(\bm{x}')_i, 0)$, using the iterative crafting process.

\textbf{ZOO} \citep{chen2017zoo} has been proposed to optimize Eq.~(\ref{eq:cw}) in the black-box manner through queries. It estimates the gradient at each coordinate as 
\begin{equation}
    \hat{g}_i = \frac{\mathcal{L}(\bm{x}+\sigma \bm{e}_i,y)-\mathcal{L}(\bm{x}-\sigma \bm{e}_i,y)}{2\sigma} \approx \frac{\partial \mathcal{L}(\bm{x},y)}{\partial x_i},
\end{equation}
where $\mathcal{L}$ is the objective in Eq.~(\ref{eq:cw}), $\sigma$ is a small constant, and $\bm{e}_i$ is the $i$-th unit basis vector. In our experiments, we perform one update with $\hat{g}_i$ at one randomly sampled coordinate. We set $\sigma=10^{-4}$ and max queries as $20,000$.

\textbf{NES} \citep{ilyas2018black} and \textbf{SPSA} \citep{uesato2018adversarial} adopt the update rule in Eq.~(\ref{eq:bim}) for adversarial example generation. Although the true gradient is unavailable, NES and SPSA give the full gradient estimation as
\begin{equation}
    \bm{\hat{g}} = \frac{1}{q}\sum_{i=1}^{q}\frac{\mathcal{J}(\bm{x}+\sigma \bm{u}_i,y)-\mathcal{J}(\bm{x}-\sigma\bm{u}_i,y)}{2\sigma}\cdot \bm{u}_i,
\end{equation}
where we use $\mathcal{J}(\bm{x}, y) = Z(\bm{x})_y - \max_{i\neq y}Z(\bm{x})_i$ instead of the cross-entropy loss, $\{\bm{u}_i\}_{i=1}^q$ are the random vectors sampled from a Gaussian distribution in NES, and a Rademacher distribution in SPSA. We set $\sigma=0.001$ and $q=128$ in our experiments, as default in the original papers.

\subsection{Adversarial training}
\label{b3}
Adversarial training (AT) is one of the most effective strategies on defending adversarial attacks, which dominates the winner solutions in recent adversarial defense competitions~\citep{brendel2020adversarial,kurakin2018competation}. The AT strategy stems from the seminal work of \citet{Goodfellow2014}, where the authors propose to craft adversarial examples with FGSM and augment them into the training data batch in a mixed manner, i.e., each mini-batch of training data consists of a mixture of clean and crafted adversarial samples. However, FGSM-based AT was shown to be vulnerable under multi-step attacks, where \citet{wong2020fast} later verify that random initialization is critical for the success of FGSM-based AT. Recent work also tries to solve the degeneration problem of one-step AT by adding regularizers~\citep{vivek2020regularizers}. Another well-known AT strategy using the mixed mini-batch manner is ALP~\citep{kannan2018adversarial}, which regularizes the distance between the clean logits and the adversarial ones. But later \citet{engstrom2018evaluating} successfully evade the models trained by ALP. As to the mixed mini-batch AT, \citet{Xie2020intriguing} show that using an auxiliary batch normalization for the adversarial part in the data batch can improve the performance of the trained models.

Among the proposed AT frameworks, the most popular one is the PGD-AT~\citep{madry2018towards}, which formulates the adversarial training procedure as a min-max problem. \citet{zhang2019theoretically} propose the TRADES framework to further enhance the model robustness by an additional regularizer between model predictions, which achieves state-of-the-art performance in the adversarial competition of NeurIPS 2018~\citep{brendel2020adversarial}. However, multi-step AT usually causes high computation burden, where training a robust model on ImageNet requires tens of GPU workers in parallel~\citep{kurakin2016adversarial,xie2018feature}. To reduce the computational cost, \citet{shafahi2019adversarial} propose the FreeAT strategy to reuse the back-propagation result for crafting the next adversarial perturbation, which facilitate training robust models on ImageNet with four GPUs running for two days.

\subsection{Metric learning + adversarial training}
\label{b4}
Previous work finds that the adversarial attack would cause the internal representation to shift
closer to the "false" class~\citep{mao2019metric,li2019improving}. Based on this observation, they propose to introduce an extra triplet loss term in the training objective to capture the stable metric space representation, formulated as
\begin{equation}
\begin{split}
        &\Loss_{\text{trip}}(z(x^*),z(x_{p}),z(x_{n}))\\
        =&\left[D(z(x^*),z(x_{p}))-D(z(x^*),z(x_{n}))+\alpha\right]^{+}\text{,}
\end{split}
\end{equation}
where $\alpha>0$ is a hyperparameter for margin, $x^*$ (anchor example) is an adversarial counterpart based on the clean input $x$, $x_{p}$ (positive example) is a clean image from the same class of $x$; $x_{n}$ (negative example) is a clean image from a different class. Here $D(u,v)$ is a distance function. \citet{mao2019metric} employ an angular distance as $D(u,v)=1-\cos{\angle(u,v)}$; \citet{li2019improving} apply the $\ell_{\infty}$ distance as $D(u,v)=\|u-v\|_{\infty}$. In the implementation, these methods apply some heuristic strategies to sample triplets, in order to alleviate high computation overhead. For example, \citet{mao2019metric} select the closest sample in a mini-batch as an approximation to the semi-hard negative example. However, the optimization on sampled triplets is still computationally expensive and could introduce class biases on unbalanced datasets~\cite{hoffer2015deep}.

\citet{zhang2019defense} apply a feature-scatter solver for the inner maximization problem of AT, which is different from PGD. Instead of crafting each adversarial example based on its clean counterpart, the feature-scatter solver generate the adversarial examples in batch to utilize inter-sample interactions, via maximizing the optimal transport (OT) distance between the clean and adversarial empirical distributions. In the implementation, they use practical OT-solvers to calculate the OT distance and maximize it w.r.t. the adversarial examples. However, the calculation of the OT distance will increase the computational burden for the AT procedure. Besides, the feature-scatter solver also leads to potential threats for the trained models to be evaded by adaptive attacks, e.g., feature attacks, as discussed before\footnote{{https://github.com/Line290/FeatureAttack}}\footnote{{https://openreview.net/forum?id=Syejj0NYvr\&noteId=rkeBhuBMjS}}.

\section{More empirical results}
In this section, we provide more empirical results and setups. In our experiments, we apply NVIDIA P100 / 2080Ti GPUs, as well as the Apex package to execute training for FastAT~\citep{coleman2017dawnbench,wong2020fast}. On CIFAR-10, all the models are trained by four GPUs in parallel for PGD-AT, ALP, and TRADES.

\subsection{Code references}
To ensure that our experiments perform fair comparison with previous work, we largely adopt the public codes and make minimal modifications on them to run the trials. Specifically, we refer to the codes of TRADES\footnote{{https://github.com/yaodongyu/TRADES}}~\citep{zhang2019theoretically}, FreeAT\footnote{{https://github.com/mahyarnajibi/FreeAdversarialTraining}}~\citep{shafahi2019adversarial}, FastAT\footnote{{https://github.com/locuslab/fast\_adversarial}}~\citep{wong2020fast} and the corrupted datasets\footnote{{https://github.com/hendrycks/robustness}} from~\citet{hendrycks2019benchmarking}. The codes are mostly based on PyTorch~\citep{paszke2019pytorch}.

\subsection{Datasets}
The CIFAR-10 dataset~\citep{Krizhevsky2012} consists of 60,000 32x32 colour images in 10 classes, with 6,000 images per class. There are 50,000 training images and 10,000 test images. We perform RandomCrop with 4 padding and RandomHorizontalFlip in training as the data augmentation. The ImageNet (ILSVRC 2012) dataset~\citep{deng2009imagenet} consists of 1.28 million training images and 50,000 validation images in 1,000 classes. As to the data augmentation, we perform RandomResizedCrop and RandomHorizontalFlip in training; Resize and CenterCrop in test. The image size is 256 and the crop size is 224.

\subsection{Extensive ablation studies}
Different choices of the scale $s$ and the margin $m$ in HE lead to different trade-offs between the clean accuracy and the adversarial robustness of the trained models, as shown in Table~\ref{appendixtable:24}. This kind of trade-off is ubiquitous w.r.t. the hyperparameter settings in different AT frameworks~\citep{kannan2018adversarial,zhang2019theoretically}.

\begin{table}[htbp]
  \vspace{-.cm}
  \caption{Classification accuracy (\%) on \textbf{CIFAR-10}. The training framework is TRADES + HE with different scale $s$ and margin $m$. We report the performance on clean inputs and under PGD-20 attack.}
  \vspace{-.1cm}
  \begin{center}
  \begin{small}
  \begin{tabular}{c|c|c|c|c}
  \hline
Defense& Scale $s$ & Margin $m$ & Clean  & PGD-20\\
\hline
\multirow{11}{*}{TRADES + HE}& 15 & 0.0 & 82.53 &60.35\\
& 15 & 0.1 & \textbf{85.00} & 61.13\\
& 15 & 0.2 & 84.88 & \textbf{62.02}\\
& 15 & 0.3 & 82.99 & 61.54\\
& 15 & 0.4 & 78.05 & 58.05\\
& 15 & 0.5 & 74.71 & 56.27\\
\cline{2-5}
& 1 & 0.2 & \textbf{89.34} & 50.33\\
& 5 & 0.2 & 85.70 & 58.75\\
& 10 & 0.2 & 85.30 & 60.17\\
& 15 & 0.2 & 84.88 & \textbf{62.02}\\
& 20 & 0.2 & 77.67 & 57.51\\
\hline
   \end{tabular}
  \end{small}
  \end{center}
  \label{appendixtable:24}
    \vspace{-.3cm}
\end{table}

\subsection{Transfer-based black-box attacks}
\begin{figure}[htbp]
\begin{center}
\vspace{-0.3cm}
\caption{Classification accuracy (\%) under the \emph{black-box} transfer-based attacks on \textbf{CIFAR-10}. The substitute models are PGD-AT, ALP and TRADES separately. * indicates white-box cases.}
\vspace{0.2cm}
\includegraphics[width=.95\columnwidth]{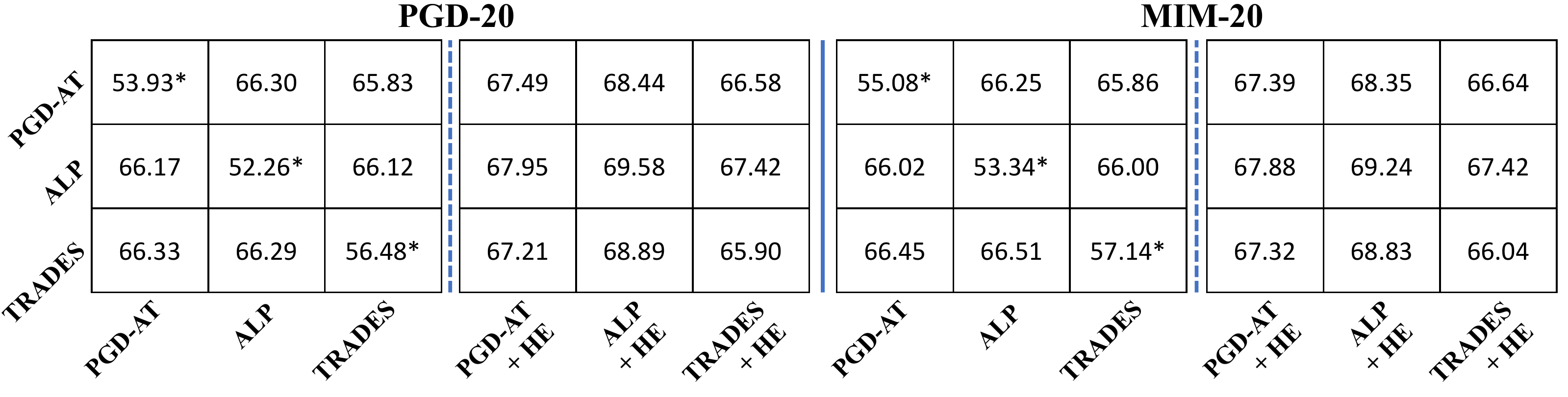}
\label{fig:4}
\end{center}
\vspace{-0.3cm}
\end{figure}

Due to the adversarial transferability~\citep{Papernot20162,Papernot2016}, the black-box adversaries can construct adversarial examples based on the substitute models and
then feed these examples to evade the original models. In our experiments, we apply PGD-AT, ALP, and TRADES to train the substitute models, respectively. To generate adversarial perturbations, we employ the untargeted PGD-20~\citep{madry2018towards} and MIM-20~\citep{Dong2017} attacks, where the MIM attack won both the targeted and untargeted attacking tracking in the adversarial competition of NeurIPS 2017~\citep{kurakin2018competation}. In Fig.~\ref{fig:4}, we show the results of transfer-based attacks against the defense models trained without or with the HE mechanism. As expected, we can see that applying HE can also better defend transfer-based attacks.

\subsection{Full results of m-HE on CIFAR-10}
In Table~\ref{appendixtable:7}, we evaluate the white-box performance of the combinations of the modified HE (m-HE) with PGD-AT, ALP, and TRADES. We set the parameters with $s=15$ and $m=0.1$. We can see that m-HE is more effective than HE when combining with PGD-AT, FreeAT and Fast AT that exclusively train on adversarial examples. In contrast, HE performs better than m-HE when combining with the frameworks training on the mixture of clean and adversarial examples, e.g., ALP and TRADES.

\begin{table*}[htbp]
\vspace{-0.3cm}
  \caption{Classification accuracy (\%) on \textbf{CIFAR-10} under the \emph{white-box} threat model. The perturbation $\epsilon=0.031$, step size $\eta=0.003$, following the setting in~\citet{zhang2019theoretically}.}
  \begin{center}
  \begin{small}
  \begin{tabular}{c|c|c|c|c|c|c|c}
  \hline
Defense& Clean & PGD-20 & PGD-500 & MIM-20 & FGSM & DeepFool & C\&W-$\ell_{\infty}$ \\
\hline
PGD-AT & \textbf{86.75} & 53.97 & 51.63 & 55.08 & 59.70 & 57.26 & 84.00\\
PGD-AT + \textbf{HE} & 86.19 & {59.36} & {57.59} & {60.19} & \textbf{63.77} & \textbf{61.56} & \textbf{84.07}\\
PGD-AT + \textbf{m-HE} & 86.25 &\textbf{59.90} &\textbf{58.46} &\textbf{60.50} &63.70 &59.47 &83.71\\
\hline

ALP & 87.18 & 52.29 & 50.13 & 53.35 & 58.99 & 59.40 & 84.96\\
ALP + \textbf{HE} & \textbf{89.91} & \textbf{57.69} & {51.78} & \textbf{58.63} & \textbf{65.08} & \textbf{65.19} & \textbf{87.86}\\
ALP + \textbf{m-HE}  &    89.23 &57.09 &\textbf{53.34} &58.04 &63.81 &60.74 &87.21\\
\hline

TRADES & 84.62 & 56.48 & 54.84 & 57.14 & 61.02 & \textbf{60.70} & 81.13\\
TRADES + \textbf{HE} & \textbf{84.88} & \textbf{62.02} & \textbf{60.75} & \textbf{62.71} & \textbf{65.69} & 60.48 & \textbf{81.44}\\
TRADES + \textbf{m-HE} & 84.30 &61.83 &60.43 &62.67 &65.49 &60.51 &80.53\\

  \hline
   \end{tabular}
  \end{small}
  \end{center}
  \label{appendixtable:7}
  \vspace{-.cm}
\end{table*}

\subsection{Full results on CIFAR-10-C and ImageNet-C}
In Table~\ref{appendixtable:7-4} and Table~\ref{appendixtable:7-3} we provide the full classification accuracy results of different defenses on CIFAR-10-C and ImageNet-C~\citep{hendrycks2019benchmarking}, respectively. These reports include detailed accuracy under $75$ combinations of severity and corruption.

\begin{table*}[htbp]
\footnotesize
\setlength{\tabcolsep}{3pt}
\vspace{0cm}
  \caption{Classification accuracy (\%) on \textbf{CIFAR-10-C}. Full results on different combination of severity and corruption. Here 'S' refers to the severity from 1 to 5, 'P' refers to PGD-AT, 'A' refers to ALP, 'T' refers to TRADES.}
  \begin{center}
  \begin{tabular}{c|c|ccc|cccc|cccc|cccc}
  \hline
\multirow{2}{*}{Defense}& \multirow{2}{*}{S} & \multicolumn{3}{c|}{\emph{Noise}}& \multicolumn{4}{c|}{\emph{Blur}}& \multicolumn{4}{c|}{\emph{Weather}}&\multicolumn{4}{c}{\emph{Digital}}\\
&  & Gauss& Shot& \!Impulse\! & \!Defocus\!& \!Glass\!& \!Motion\!& \!Zoom\! &Snow & Frost & Fog & Bright &\!Contra\! & \!Elastic\!& Pixel& \!JPEG\!\\
\hline
\multirow{5}{*}{P} & 1& 85.97 & 86.3& 83.55& 86.15 & 81.92& 83.65& 82.86 & 86.12& 84.88& 84.27 & 87.15& 82.37& 82.06 & 86.23 & 85.17 \\
& 2&  84.16 &  85.82&  79.92&  84.8 &  82.06&  80.09&  82.49 &  84.4&  80.84&  76.22 &  86.68&  59.77&  82.2 &  85.62 &  84.64 \\
& 3&  81.22 &  83.08&  76.79&  82.87 &  81.55&  76.32&  81.12 &  82.39&  75.24&  65.71 &  85.7&  40.64&  81.13 &  85.33 &  84.41 \\
& 4&  79.4 &  81.2&  69.55&  80.67 &  76.18&  76.39&  80.17 &  78.03&  75.93&  51.96 &  83.87&  23.09&  79.65 &  84.27 &  83.86 \\ 
& 5&  77.48 &  78.02&  62.8&  74.7 &  76.72&  71.66&  77.77 &  75.69&  72.9&  30.37 &  76.86&  16.9&  78.95 &  82.37 &  83.69 \\

\hline
\multirow{5}{*}{P \!+ \!\textbf{HE}} & 1& 85.37 & 85.67& 83.7& 85.63 & 81.51& 83.75& 82.67 & 85.33& 84.26& 84.1 & 86.45& 82.43& 81.69 & 85.43 & 85.05 \\
& 2&  83.69 &  84.99&  80.97&  84.5 &  81.76&  80.66&  82.33 &  83.99&  80.33&  76.84 &  86.0&  61.36&  81.92 &  85.15 &  84.52 \\
& 3&  81.14 &  82.46&  78.01&  82.65 &  81.2&  76.73&  81.11 &  81.43&  75.71&  66.59 &  85.17&  41.76&  81.2 &  84.63 &  84.05 \\
& 4&  79.74 &  81.25&  71.67&  80.83 &  76.14&  77.03&  80.09 &  77.77&  76.35&  52.75 &  83.31&  23.65&  80.01 &  83.77 &  83.67 \\
& 5&  77.76 &  77.99&  66.37&  75.58 &  76.67&  72.68&  78.16 &  75.35&  73.28&  31.96 &  77.44&  16.36&  78.61 &  81.84 &  83.23 \\

\hline
\multirow{5}{*}{A} & 1& 86.57 & 86.93& 84.13& 86.53 & 83.04& 84.12& 83.07 & 86.4& 85.77& 85.12 & 87.4& 83.21& 82.18 & 86.46 & 85.79 \\
& 2&  84.96 &  86.31&  80.48&  85.01 &  82.65&  81.04&  82.73 &  84.86&  82.19&  77.27 &  87.24&  61.34&  82.11 &  86.02 &  85.09 \\
& 3&  81.84 &  83.55&  77.23&  82.84 &  81.83&  76.91&  81.15 &  82.76&  76.57&  65.68 &  86.36&  41.96&  81.32 &  85.73 &  84.63 \\
& 4&  80.05 &  81.64&  69.53&  80.65 &  77.1&  77.09&  80.17 &  78.56&  77.54&  51.09 &  84.74&  24.92&  80.13 &  84.46 &  84.23 \\
& 5&  78.01 &  78.6&  62.97&  74.69 &  76.96&  71.99&  77.75 &  76.13&  74.26&  28.39 &  78.66&  17.85&  78.81 &  82.79 &  83.68 \\

\hline
\multirow{5}{*}{A \!+ \!\textbf{HE}} & 1& 88.61 & 89.27& 85.56& 89.58 & 83.58& 87.05& 86.36 & 88.71& 88.62& 88.2 & 90.1& 86.55& 85.93 & 89.37 & 88.52 \\
& 2&  85.89 &  88.25&  81.13&  88.15 &  83.79&  83.83&  85.88 &  87.0&  86.2&  82.37 &  89.9&  69.33&  85.97 &  88.72 &  87.62 \\
& 3&  81.61 &  83.99&  76.29&  86.14 &  83.69&  79.81&  84.68 &  85.71&  82.47&  73.58 &  89.13&  51.2&  84.94 &  87.98 &  86.99 \\
& 4&  78.72 &  81.42&  67.61&  84.08 &  75.73&  80.34&  83.73 &  82.52&  82.79&  61.5 &  88.06&  30.06&  83.36 &  86.92 &  86.89 \\
& 5&  76.26 &  76.64&  60.84&  78.21 &  77.56&  75.28&  81.5 &  81.76&  79.36&  38.48 &  84.46&  16.58&  82.01 &  84.19 &  86.37 \\

\hline 
\multirow{5}{*}{T} & 1& 83.74 & 84.16& 81.61& 84.01 & 79.97& 81.98& 80.69 & 84.16& 83.57& 82.53 & 85.21& 79.91& 79.47 & 84.22 & 83.27 \\
& 2&  81.84 &  83.44&  78.34&  82.61 &  79.85&  78.62&  80.04 &  82.96&  79.7&  74.42 &  84.78&  57.63&  79.8 &  83.45 &  82.64 \\
& 3&  78.63 &  80.56&  74.84&  80.58 &  79.55&  75.09&  78.9 &  80.79&  73.97&  63.06 &  83.78&  39.34&  79.05 &  83.07 &  82.32 \\
& 4&  77.09 &  78.42&  67.85&  78.62 &  74.7&  75.0&  77.9 &  76.33&  75.08&  49.91 &  82.13&  24.6&  77.48 &  82.2 &  81.83 \\
& 5&  74.8 &  75.27&  61.88&  73.4 &  74.55&  71.03&  75.78 &  73.39&  72.41&  28.5 &  74.41&  17.54&  76.86 &  80.31 &  81.53 \\

\hline
\multirow{5}{*}{T \!+ \!\textbf{HE}} & 1& 83.06 & 83.78& 81.12& 83.96 & 79.42& 82.13& 81.28 & 83.69& 83.22& 82.31 & 85.0& 80.2& 80.35 & 83.84 & 83.24 \\
& 2&  81.18 &  83.02&  78.41&  82.89 &  79.75&  79.39&  81.03 &  82.23&  79.21&  75.12 &  84.93&  60.38&  80.14 &  83.23 &  82.64 \\
& 3&  78.43 &  80.11&  75.4&  81.35 &  79.53&  76.19&  79.9 &  80.13&  73.93&  65.6 &  83.87&  41.97&  79.88 &  83.12 &  82.37 \\
& 4&  76.85 &  78.63&  69.59&  79.78 &  74.03&  76.62&  78.95 &  76.34&  74.51&  52.36 &  82.2&  25.59&  78.33 &  82.16 &  82.07 \\
& 5&  74.89 &  75.55&  64.55&  74.77 &  75.32&  71.98&  76.92 &  73.74&  71.79&  31.56 &  75.64&  17.07&  77.74 &  80.17 &  81.66 \\

\hline

  \hline
   \end{tabular}
  \end{center}
  \label{appendixtable:7-4}
  \vspace{0cm}
\end{table*}

\begin{table*}[htbpt]
\footnotesize
\setlength{\tabcolsep}{3pt}
\vspace{0cm}
  \caption{Classification accuracy (\%) on \textbf{ImageNet-C}. Full results on different combination of severity and corruption. Here 'S' refers to the severity from 1 to 5, 'F' refers to FreeAT.}
  \begin{center}
  \begin{tabular}{c|c|ccc|cccc|cccc|cccc}
  \hline
\multirow{2}{*}{Defense}& \multirow{2}{*}{S} & \multicolumn{3}{c|}{\emph{Noise}}& \multicolumn{4}{c|}{\emph{Blur}}& \multicolumn{4}{c|}{\emph{Weather}}&\multicolumn{4}{c}{\emph{Digital}}\\ 
&  & Gauss& Shot& \!Impulse\! & \!Defocus\!& \!Glass\!& \!Motion\!& \!Zoom\! &Snow & Frost & Fog & Bright &\!Contra\! & \!Elastic\!& Pixel& \!JPEG\!\\
\hline
\multirow{5}{*}{F} & 1 & 54.26& 53.23& 44.00& 33.91& 42.25& 43.97& 38.64& 42.57& 45.83& 11.00& 56.42& 18.29& 48.45& 53.91& 54.74\\

& 2 & 45.84& 43.08& 32.60& 26.68& 34.06& 34.70& 33.36& 26.16& 28.50& 4.41& 53.20& 6.59& 31.78& 52.98& 53.92\\

& 3 & 29.74& 28.36& 23.48& 16.82& 25.22& 23.83& 26.93& 22.69& 17.22& 1.54& 47.98& 1.38& 49.64& 50.19& 53.34\\

& 4 & 12.96& 10.66& 8.91& 11.10& 19.35& 15.20& 23.52& 11.98& 15.51& 1.20& 39.56& 0.40& 46.09& 45.24& 51.57\\

& 5 & 3.28& 4.85& 2.73& 7.25& 12.29& 11.05& 18.84& 11.75& 10.28& 0.43& 28.73& 0.34& 32.82& 41.55& 49.19\\
\hline
\multirow{5}{*}{F \!+ \!\textbf{HE}} & 1 &55.14& 53.27& 44.29& 38.15& 46.08& 47.57& 41.90& 45.92& 50.32& 15.25& 58.73& 23.31& 51.15& 56.17& 57.03\\

& 2 & 43.96& 40.02& 29.08& 30.34& 37.97& 38.39& 36.21& 30.21& 34.26& 6.59& 56.47& 9.23& 34.61& 55.40& 56.23\\

& 3 & 24.49& 23.14& 18.38& 18.52& 27.98& 26.44& 29.38& 26.83& 22.34& 2.39& 52.39& 1.71& 52.64& 52.65& 55.77\\

& 4 & 9.37& 8.37& 5.89& 11.71& 21.45& 16.42& 25.61& 15.22& 20.43& 1.85& 45.56& 0.47& 48.94& 47.82& 54.17\\

& 5 & 2.23& 3.95& 1.63& 7.09& 12.93& 11.59& 20.68& 14.89& 14.36& 0.58& 36.21& 0.41& 36.25& 43.81& 52.04\\
\hline

  \hline
   \end{tabular}
  \end{center}
  \label{appendixtable:7-3}
  \vspace{0cm}
\end{table*}


\end{document}